\pdfoutput=1

\documentclass[11pt]{article}

\usepackage[final]{acl}
\usepackage{enumerate}
\usepackage{times}
\usepackage{latexsym}
\usepackage{hyperref}
\usepackage[T1]{fontenc}

\usepackage[utf8]{inputenc}

\usepackage{microtype}

\usepackage{inconsolata}

\usepackage{graphicx}

\usepackage{graphicx}
\usepackage{enumitem}
\usepackage{bbding}
\usepackage{color}
\usepackage[noend]{algpseudocode}
\usepackage{algorithmicx,algorithm}
\usepackage{booktabs}
\usepackage{threeparttable}
\usepackage{multirow}
\usepackage{amsmath}
\usepackage{url}
\usepackage{tikz}
\usepackage{easyReview}

\usepackage[normalem]{ulem}

\title{Beyond Persuasion: Towards Conversational Recommender System \\ with Credible Explanations}

\author{
Peixin Qin$^{\spadesuit\heartsuit}$, \quad
Chen Huang$^{\spadesuit\heartsuit}$, \quad
Yang Deng$^{\clubsuit}$, 
\quad 
\textbf{Wenqiang Lei}$^{\spadesuit\heartsuit}$\thanks{Corresponding author.}, \quad
\textbf{Tat-Seng Chua}$^{\diamondsuit}$ \quad
\\
${\spadesuit}$ Sichuan University \quad ${\clubsuit}$ Singapore Management University \quad \\
${\heartsuit}$ Engineering Research Center of Machine Learning and Industry Intelligence, \\ Ministry of Education, China \\
${\diamondsuit}$ National University of Singapore \\
\{qinpeixin.scu, huangc.scu\}@gmail.com \quad ydeng@smu.edu.sg \\ \{wenqianglei, lvjiancheng\}@scu.edu.cn \quad chuats@comp.nus.edu.sg
}

\begin{document}
\maketitle
\begin{abstract}
With the aid of large language models, current conversational recommender system (CRS) has gaining strong abilities to persuade users to accept recommended items. While these CRSs are highly persuasive, they can mislead users by incorporating incredible information in their explanations, ultimately damaging the long-term trust between users and the CRS. To address this, we propose a simple yet effective method, called PC-CRS, to enhance the credibility of CRS's explanations during persuasion. It guides the explanation generation through our proposed credibility-aware persuasive strategies and then gradually refines explanations via post-hoc self-reflection. Experimental results demonstrate the efficacy of PC-CRS in promoting persuasive and credible explanations. Further analysis reveals the reason behind current methods producing incredible explanations and the potential of credible explanations to improve recommendation accuracy.
\end{abstract}

\section{Introduction}

Conversational Recommender Systems (CRSs) aims to engage in a natural language conversation with users, provide recommendations, and ultimately achieve a high level of user acceptance \cite{jannach2021survey, gao2021advances}. 
To achieve this, providing proper recommendation explanations along with accurate recommendations is paramount, because users are usually not familiar with the recommendations \cite{ijcai20-exp-crs,sigir23-exp-crs}.
Such explanations should be carefully crafted, incorporating persuasive elements that can influence user behavior and decision-making, thus increasing the likelihood of user acceptance of the recommendations \cite{alslaity2019towards, yu2011toward}.
Recently, the integration of Large Language Models (LLMs) has dramatically enhanced the persuasive power of current CRSs. LLMs possess the remarkable ability to generate highly convincing content that can rival, and even surpass, human-crafted persuasion \cite{hackenburg2023comparing, carrasco2024large}, significantly augmenting CRSs in delivering persuasive explanations, which improve user understanding and ultimately result in higher acceptance rates \cite{huang2024concept}. 

\begin{figure}[t]
    \centering
    \setlength{\abovecaptionskip}{5pt}   
    \setlength{\belowcaptionskip}{2pt}
    \includegraphics[width=0.49\textwidth]{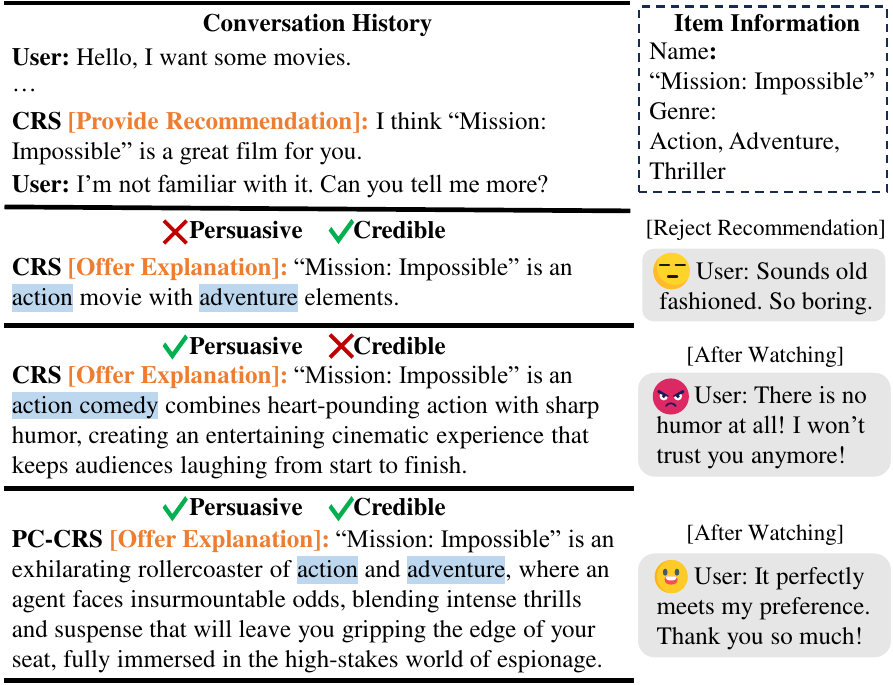}
    \caption{Examples of persuasive and credible explanations. A persuasive and credible CRS would eventually foster the long-term trust to users.}
    \label{fig:1}
    \vspace{-5mm}
\end{figure}

While LLM-based CRS is highly persuasive, a concerning trend has been observed: these CRSs can mislead users by incorporating deceptive elements into their explanations. For example, as illustrated in Figure \ref{fig:1}, a CRS mistakenly recommends the film "\textit{Mission: Impossible}" as a comedy movie to unfamiliar users, ultimately resulting in a negative user experience after viewing the film. This practice contradicts the formal definition of persuasion, which emphasizes influencing people's behaviors or attitudes \textbf{without using coercion or deception} \cite{reardon1991persuasion, oinas2008systematic}. Such incredible explanations can create erroneous perceptions about recommended items \cite{adomavicius2013recommender}, ultimately damaging the long-term trust between users and the CRS  \cite{koranteng2023credibility, deng2022user, angell1990managing}. While the need to enhance credibility during persuasion in CRSs is acknowledged \cite{huang2024concept}, effective solutions remain elusive.

To this end, we introduce a simple yet effective method for \underline{P}ersuasive and \underline{C}redible \underline{CRS}, called \textbf{PC-CRS}. 
It proactively emphasizes both persuasiveness and credibility during the generation of explanations, and then gradually refines them via post-hoc self-reflection.
This is achieved by a two-stage process: Strategy-guided Explanation Generation and Iterative Explanation Refinement. Specifically, in the first stage, PC-CRS utilizes the novel Credibility-aware Persuasive Strategies to guide the generation of candidate explanations. Such strategies are informed by social science research on persuasion \cite{fogg2002persuasive, cialdini2004social} and further tailored with credible information to ensure both persuasive and credible explanations in our scenario.
In the second stage, PC-CRS utilizes a Self-Reflective Refiner to identify and correct potential misinformation in the candidate explanations. 
It is due to generative models have the inherent tendency to prioritize contextual coherence at the expense of faithful adherence to source information \cite{miao2021prevent, chen2022towards, chen2023beyond}.
As such, PC-CRS prevents potential deception in candidates, thus enhancing credibility. Its training-free nature also makes it a highly efficient and adaptable solution.

We conduct extensive experiments to demonstrate the effectiveness of PC-CRS. Our experiments leverage the widely-used simulator-based evaluation framework\footnote{Human evaluation in Section \ref{sec:ablation} validates our reliability.} \cite{wang2023rethinking, huang2024concept} and employ two CRS benchmarks: Redial \cite{li2018towards} and OpendialKG \cite{moon2019opendialkg}. Experimental results show that PC-CRS, on average, achieves an improvement of 8.17\% on credibility score (i.e., consistency with factual information) and 5.07\% on persuasiveness score (i.e., raising user's watching intention towards recommended items), compared to the best baseline. 
Further analysis reveals the reason why LLM-based CRS generates incredible explanations is that they cater to user's history utterances rather than describing items faithfully.
In addition, the in-depth analysis also suggests that our credible explanations promote recommendation accuracy. 
This is potentially due to that credible explanations avoid the introduction of noisy information and contribute to a reliable conversation context, making it easier to comprehend user's true preference.
Our main contributions are as follows:
\begin{itemize}[leftmargin=*, itemsep=-4pt]
    \item For the first time, we investigate the crucial role of bolstering credibility during CRS persuasion, which fosters the long-term trust to users.
    \item We propose a novel method, PC-CRS, for generating both persuasive and credible recommendation explanations, with credibility-aware persuasive strategies and self-reflective refinement.
    \item We conduct extensive experiments to validate the effectiveness of PC-CRS in both persuasiveness and credibility. In-depth analysis reveals the reason behind current methods producing incredible explanations and the potential of credible explanations for improving recommendation accuracy.
\end{itemize}

\begin{figure*}[t]
    \centering
    \setlength{\abovecaptionskip}{5pt}   
    \setlength{\belowcaptionskip}{2pt}
    \includegraphics[width=1\textwidth]{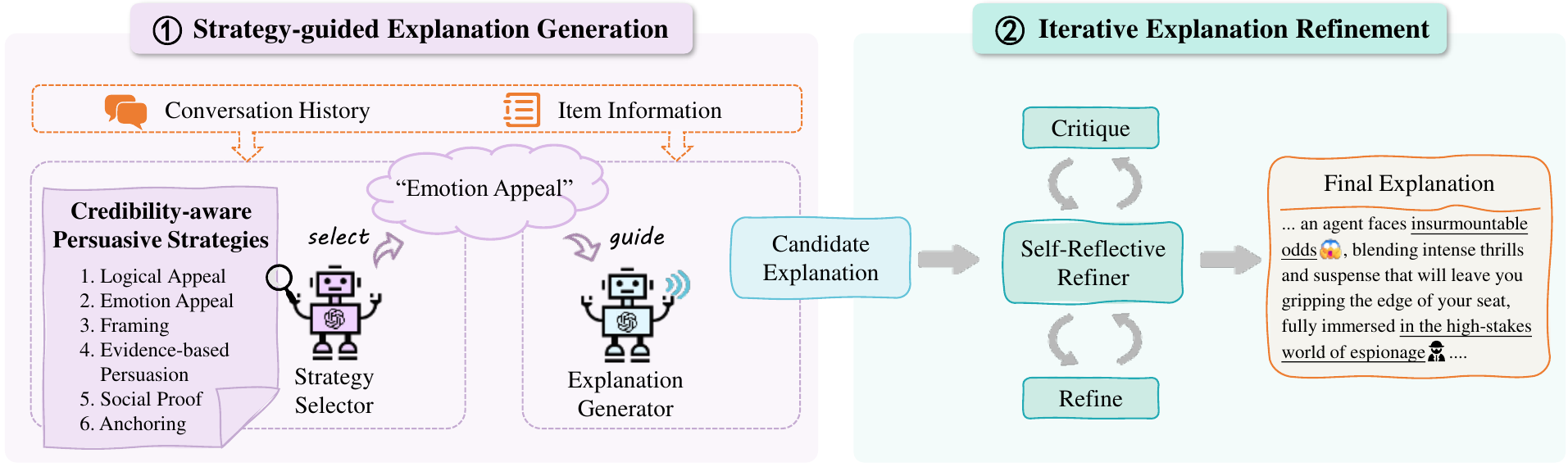}
    \caption{Two-stage process of PC-CRS. It first selects an appropriate strategy that is used to generate a candidate explanation. Then, Self-Reflective Refiner eliminates the misinformation in the candidate in an iterative way.}
    \label{fig:2}
    \vspace{-5mm}
\end{figure*}

\section{Related Work}

Our research focuses on the explanations of CRSs, particularly highlighting their persuasiveness and credibility. Hence, we provide an overview of CRS and persuasive and credible recommender systems and then discuss our differences.  


\textbf{CRS.} 
CRS enables users to engage in free-form natural language conversations with the system to achieve their recommendation-related goals \cite{li2018towards, deng2021unified, deng2022leveraging, deng2023unified, li2024incorporating}.
To generate human-like responses, early studies leverage pre-trained language models (PLMs) as their backbones \cite{aacl22-crs,wang2022barcor, wang2022towards}, enabling them to proactively interact and engage with users through verbal explanations \cite{ijcai20-exp-crs, sigir23-exp-crs, zhang2024clamberbenchmarkidentifyingclarifying}.
In the era of LLMs, CRSs have shifted from providing simple information to actively persuading users during explanations \cite{wang2023rethinking}, ultimately increasing user acceptance \cite{yu2011toward, alslaity2019towards}. 
While LLMs enable CRSs to generate highly persuasive explanations, a recent study revealed a concerning trend: they may incorporate misinformation to achieve persuasiveness \cite{huang2024concept}, jeopardizing the long-term relationship of trust between users and the CRS.
To address this challenge, we propose a method to enhance the credibility of CRS explanations during persuasion.

\textbf{Persuasive and Credible Recommender Systems.}
Early research on identifying how people persuade others with credible explanations in recommendations draws heavily on insights from social science and human-computer interaction \cite{fogg1999elements, cialdini2001science, fogg2002persuasive, deng2024towards}.
These findings resonate with human studies on recommender systems, which consistently showed that users are more inclined to accept recommendations from sources perceived as persuasive and credible \cite{o2002guilt, gkika2014persuasive}. Besides these human studies, theoretical frameworks on enhancing the persuasiveness \cite{oinas2008systematic, alslaity2019towards, slattery2020persuasion} or credibility \cite{yoo2006measuring, yoo2010creating} of a recommender system are also proposed. 
However, the main limitation of these works is that they focus on exploring the feasibility of using persuasive or credible features through theoretical analysis or human studies, rather than designing practical methods.
We address this gap by introducing PC-CRS in practice and conducting empirical studies to show its effectiveness.


\section{PC-CRS}

\textbf{Overview}. Our PC-CRS, as illustrated in Figure \ref{fig:2}, involves a two-stage process, i.e., Strategy-guided Explanation Generation and Iterative Explanation Refinement. 
Given the conversation history and item information, the former stage selects an appropriate strategy from Credibility-aware Persuasive Strategies and then generates a candidate explanation accordingly. Then, taking the previous candidate as input, the latter stage critiques and refines it to eliminate misinformation and yield the final explanation. 
PC-CRS leverages LLMs with detailed Chain-of-Thought instructions \cite{kojima2022large} to make full use of the generative capabilities of them in the above two stages.


\subsection{Strategy-guided Explanation Generation}
As previous CRSs often lack of explicit focus on persuasiveness and credibility, this stage aims to proactively emphasize the two factors when offering explanations in PC-CRS.
To achieve this, we take inspirations from social science research \cite{cialdini2004social, fogg2002persuasive, zeng2024johnny} and tailor them to develop our Credibility-aware Persuasive Strategies, guiding the explanation generation process of PC-CRS.


\subsubsection{Credibility-aware Persuasive Strategies}

Drawing upon the well-established Elaboration Likelihood Model of persuasion \cite{cacioppo1986central}, we propose six credibility-aware persuasive strategies specifically tailored for credible CRS that encourage the use of factual information during persuasion. These strategies are categorized into three groups. In particular, the first three strategies aim to persuade individuals with carefully constructed content, while the next two aim to influence users through peripheral cues (e.g., the source's credibility), and the last one combines elements of both. We further specify the credible information used in these strategies and construct prompts for the LLM to effectively use these strategies. Strategy examples and detailed prompts are shown in Appendix \ref{sec:appendix} and Appendix \ref{app:prompt}, respectively. 
\begin{itemize}[leftmargin=*, itemsep=-4pt]
    \item \textbf{Logical Appeal (L.A.)} refers to faithfully presenting the logic and reasoning process of the system to influence people \cite{cronkhite1964logic}, e.g, describing how a movie's genre is consistent with user's preference.
    By this means, users can see "why" a particular recommendation is suggested and know the "subjectivity" of machine's logic, leading them to trust and accept the recommendations. 
    \item \textbf{Emotion Appeal (E.A.)} refers to eliciting specific emotions and sharing credible and impactful stories to foster trust and deep connection with users \cite{petty2003emotional}, e.g., sharing a movie's plot to elicit user's emotion. Validating users' feelings through the system's explanations can build credibility by breaking down barriers and making it easier to influence user's decisions.
    \item \textbf{Framing (Fr.)} refers to emphasizing the positive aspects or outcomes of a decision in a trustworthy manner \cite{perloff1993dynamics}, e.g., highlighting the positive experience of watching the movie. This strategy honestly enhances the perceived benefits of a decision, making the recommendation more appealing and attractive.
    \item \textbf{Evidence-based Persuasion (E.P.)} refers to using empirical data or objective and verifiable facts to support a claim or decision \cite{o2016evidence}, e.g., showing awards of a movie. This strategy reduces the influence of biases and subjective opinions by showing objective information in real world, making it both credible and convincing.
    \item \textbf{Social Proof (S.P.)} refers to emphasizing the behaviors or endorsements of the majority in real world to support claims \cite{cialdini2004social}, e.g., presenting a movie's rating or reviews. This technique originates from the subjectivity of other users and leverages the psychological tendency of individuals to conform to the actions or beliefs of others, thereby increasing the persuasive impact and credibility of explanations.
    \item \textbf{Anchoring (An.)} refers to relying on an initial, credible piece of information as a reference point to gradually influence or persuade the user \cite{cialdini2004social}, e.g., first showing a movie's awards to attract users and then describing its genre and plot. People rely on the first piece of information they receive to make decisions. If this anchor is credible, it builds trust and influences subsequent decisions, making the persuasion more effective.
\end{itemize}

With the guidance of these strategies, PC-CRS is capable of increasing its awareness of generating persuasive and credible explanations. 

\subsubsection{Explanation Generation}
As the conversation proceeds, we select suitable strategy to guide the explanation generation at each turn, which helps adapt to the dynamics of dialogue contexts \cite{wang2019persuasion}.
As shown in Figure \ref{fig:2}, PC-CRS prompts the LLM with detailed instructions to select a strategy and generate an explanation candidate accordingly.

\textbf{Strategy Selection.}
Given a recommended item, PC-CRS retrieves its detailed information from a credible source (e.g., a knowledge base). Then, taking the conversation history $H$ and retrieved item information $I$ as inputs, a strategy selector, powered by the LLM, chooses an appropriate strategy $s$ from Credibility-aware Persuasive Strategies $S$: 
\begin{equation}
    s = StrategySelector(H, I, S).
\end{equation}

\textbf{Explanation Candidate Generation.} Given the selected strategy $s$, the conversation history $H$, and item information $I$, we prompt the LLM to produce recommendation explanation candidates:
\begin{equation}
    c = ExplanationGenerator(H, I, s).
\end{equation}
As such, PC-CRS customizes explanation candidates to match the user's preferences and context, making the interaction more relevant and engaging. Besides, the Credibility-aware Persuasive Strategies also explicitly guide the explanations to be both persuasive and credible.

\subsection{Iterative Explanation Refinement}
As generative models tend to prioritize contextual coherence at the expense of faithfully adhering to the source information \cite{miao2021prevent, chen2022towards}, there still might be illusory details incorporated in the candidate explanations. To this end, PC-CRS aims to analyze the factual basis and plausibility of each claim, ultimately ensuring that only credible and well-supported explanations are presented to the user. To achieve this, this stage, inspired by the self-reflection mechanism \cite{ji2023towards, madaan2024self}, leverages a self-reflective refiner to criticize and refine incredible claims within the candidates iteratively. 

\textbf{Critique.} Each explanation candidate is treated as an initial proposal. In the $k$-th iteration, a critic examines if the candidate explanation $c_k$ contains any misinformation based on item information $I$:
\begin{equation}
    {cq}_k = Critic(c_k, I).
\end{equation}
The critic utilizes a self-reflective approach, starting by summarizing the claims within the explanation candidate. This summary is then compared against the relevant item information. Operating independently of any conversational context, the critic generates a critique (${cq}_k$) which evaluates the explanation's credibility. This critique identifies whether further refinement is necessary and, if so, suggests specific improvements.

\textbf{Refinement.} If the critic deems refinement necessary, the refiner plays a vital role in generating a revised explanation. This refinement process leverages both the original explanation ($c_k$) and the critic's feedback (${cq}_k$) to produce the new one:
\begin{equation}
    c_{k+1} = Refiner(H, I, s, c_k, {cq}_k).
\end{equation}
Here, the refiner is tasked with removing misinformation from the candidate while maintaining consistency with the conversation history and selected strategy. This process cycles between critique and refinement steps, continuing until a preassigned stopping condition is met. This condition is triggered either when the critic indicates that no further refinement is necessary or after reaching the maximum number of iterations (2 in our practice).

As such, PC-CRS gradually eliminates misinformation in the candidate and outputs a final explanation that is both persuasive and credible. PC-CRS achieves this process in a training-free manner, making it an efficient and adaptable solution.


\section{Experiments}
In this section, we investigate the superiority of PC-CRS on persuasiveness and credibility (Section \ref{sec:main}). Subsequently, we provide detailed analyses on characteristics of PC-CRS to gain understanding why it alleviates the incredibility and improves the recommendation accuracy (Section \ref{sec:analysis}).
Then, ablation studies and human evaluation indicate the necessity of two stages in PC-CRS and the reliability of our evaluation, respectively (Section \ref{sec:ablation}).

\begin{table*}[t]
\centering
\setlength{\abovecaptionskip}{5pt}   
\setlength{\belowcaptionskip}{2pt}
\resizebox{0.95\textwidth}{!}{
\begin{tabular}{cc|ccc|ccc}
\toprule
\multicolumn{2}{c|}{\multirow{3}{*}{Models}}                      & \multicolumn{3}{c|}{Redial}                                                                                                      & \multicolumn{3}{c}{OpendialKG}                                                                                                   \\ \cline{3-8} 
\multicolumn{2}{c|}{}                                             & Persuasiveness                 & Credibility                   & \begin{tabular}[c]{@{}c@{}}Convincing\\ Acceptance\end{tabular} & Persuasiveness                 & Credibility                   & \begin{tabular}[c]{@{}c@{}}Convincing\\ Acceptance\end{tabular} \\ \midrule
\multicolumn{1}{c|}{\multirow{2}{*}{PLM-based}} & BARCOR          & 34.44                          & 2.23                          & /                                                               & 20.27                          & 1.95                          & /                                                               \\
\multicolumn{1}{c|}{}                           & UniCRS          & 13.74                          & 2.77                          & /                                                               & 25.57                          & 2.42                          & /                                                               \\ \hline
\multicolumn{1}{c|}{\multirow{4}{*}{LLM-based}} & InterCRS         & 73.05                          & 3.50                          & 63.01                                                           & 76.36                          & \underline{3.85} & \underline{71.30}                                  \\
\multicolumn{1}{c|}{}                           & ChatCRS         & 71.68                          & 3.66                          & \underline{73.89}                                 & \underline{79.64} & 3.26                          & 66.67                                                           \\
\multicolumn{1}{c|}{}                           & MACRS           & \underline{76.77} & \underline{3.87} & 73.86                                                           & 78.89                          & 3.14                          & 59.34                                                           \\
\multicolumn{1}{c|}{}                           & PC-CRS (\textit{ours}) & \textbf{82.12}                 & \textbf{4.15}                 & \textbf{78.07}                                                  & \textbf{82.16}                 & \textbf{4.20}                 & \textbf{87.67}                                                  \\ \hline
\multicolumn{2}{c|}{\textbf{Improvement} (\%)}                              & 6.97$\uparrow$    & 7.24$\uparrow$   & 5.66$\uparrow$                                     & 3.16$\uparrow$    & 9.10$\uparrow$   & 22.96$\uparrow$                                    \\ \bottomrule
\end{tabular}}
\caption{Results in terms of persuasiveness and credibility. We report our improvement to the best baseline (\underline{underlined}). LLM-based CRSs suffer from the incredibility issue during persuasion. PC-CRS generates credible and persuasive explanations. PLM-based CRSs have no user acceptance thus Convincing Acceptance is incalculable.}
\label{tab:main}
\vspace{-5mm}
\end{table*}

\subsection{Experimental Setup}
\label{setting}
\textbf{User Simulator \& Datasets.}
Utilizing a user simulator to evaluate CRS is a common practice, as interacting with real humans can be quite expensive \cite{lei2020estimation, wang2023rethinking, wang2023improving, fang2024multi}.
In accordance with prior research, we follow the simulator in \citet{wang2023rethinking, huang2024concept} tailored for two CRS benchmarks, namely Redial \cite{li2018towards} and OpendialKG \cite{moon2019opendialkg}.
Specifically, the simulator is initialized with different user preferences and personas. To mimic real-world scenarios, it has only access to a combination of preferred attributes without certain target items.
During the conversation, the CRS and simulator converse with each other in free-form natural language. The conversation ends either when the maximum number of turns is reached or the simulator accepts the recommendation provided by the CRS. See more details on the simulator and datasets in Appendix \ref{appendix:simulator}.

\textbf{Baselines.} 
We compare PC-CRS with SOTA \underline{PLM-based methods}, i.e., BARCOR \cite{wang2022barcor} and UniCRS \cite{wang2022towards}. We also compare PC-CRS with recent \underline{LLM-based CRSs}, including InterCRS \cite{wang2023rethinking}, ChatCRS \cite{li2024incorporating} and MACRS \cite{fang2024multi}. 

\textbf{Evaluation Metrics.} 
Following \citet{ye2023flask, liu2023calibrating}, we utilize the GPT-4-based evaluator, equipped with fine-grained scoring rubrics, to achieve a cost-effective evaluation (we also involve human evaluation in Section \ref{sec:ablation}). Concretely, we introduce three metrics to quantitatively measure the performance of CRS explanations:
\begin{itemize}[leftmargin=*, itemsep=-3pt]
    \item \underline{Persuasiveness}. Inspired by human studies on persuasion \cite{lu2023user}, Persuasiveness score focuses on to what extent an explanation can change the watching intention of a user towards the recommended item.
    This is achieved by instructing the evaluator to score its watching intention, ranging from 1 to 5. Specifically, the evaluator rates its initial intention $i_{pre}$ based solely on the item's title. Then it is required to rate the intention $i_{post}$ after reading the CRS explanation. Finally, the evaluator rates the 'true' intention $i_{true}$ after seeing the full information about the item.
    And the Persuasiveness is calculated as follows. A higher Persuasiveness score means a stronger ability in arousing user's watching intention towards recommended items. 
    \begin{equation}
        Persuasiveness = 1 - \frac{i_{true} - i_{post}}{i_{true} - i_{pre}}.
    \end{equation}

    \item \underline{Credibility}. We resort to metrics used in text summarization to access utterance-level credibility, checking if each explanation (summary) is consistent with the facts (source texts). Following \citet{gao2023human, luo2023chatgpt}, we employ GPT-4 and prompt it to score the Credibility ranging from 1 to 5 with a detailed criteria\footnote{In the following, we call the Credibility score less than 3 as \textit{low credibility} and greater than 3 as \textit{high credibility}.}.

    \item \underline{Convincing Acceptance}. This metric aims to assess dialogue-level credibility.
    It measures how often the CRS successfully convinces the simulator to accept a recommendation while maintaining a high credibility.
    A higher Convincing Acceptance indicates a lower likelihood of users being misled by deceptive explanations.
\end{itemize}
In addition to evaluating the quality of CRS explanations, we also employ metrics to evaluate the recommendation accuracy of CRSs. Following \citet{wang2023rethinking} and \citet{zhang2023variational}, we use \underline{Success Rate} (SR) and \underline{Recall@$k$} (R@$k$), where $k=1,5,10$.

\textbf{Implementation Details.} All baselines are implemented by checkpoints and prompts from the corresponding code repositories or papers. For a fair comparison, all LLM-based CRSs including PC-CRS employ the dual-tower encoder \cite{neelakantan2022text} as the recommendation module to retrieve items.
Following previous LLM-based CRS, we employ ChatGPT\footnote{gpt-3.5-turbo-0125} to implement the user simulator and PC-CRS. 
Additionally, GPT-4\footnote{gpt-4o-2024-05-13} is employed as the evaluator due to its advanced ability in evaluating natural language generation tasks \cite{liu2023g}. Details on implementations and prompts are provided in Appendix \ref{appendix:prompts}\footnote{\href{https://github.com/mumen798/PC-CRS}{https://github.com/mumen798/PC-CRS}}.

\subsection{Main Results}
\label{sec:main}

We start by examining whether PC-CRS achieves the goal of enhancing credibility during persuasion.
Table \ref{tab:main} shows the performance of PC-CRS and other baselines. We also conduct experiments on PC-CRS using Llama3-8B-instruct as the backbone to demonstrate that our PC-CRS can generalize to various LLM options (see Appendix \ref{appendix:experiment} for details). Here, our findings are as follows.


\textbf{LLM-based CRSs are highly persuasive.}
As shown in Table \ref{tab:main}, LLM-based systems achieve a Persuasiveness score that is 3.3 times higher than their PLM-based counterparts on average. This superior performance is attributed to the LLMs' inherent strength in comprehending user needs and effectively modeling context, leading to more convincing and impactful recommendations.


\textbf{PC-CRS achieves both persuasive and credible explanations.} 
According to Table \ref{tab:main}, we observed that PLM-based CRSs struggle to generate credible or persuasive explanations, resulting in no recommendations being accepted. This limitation stems from their relatively weak generation capabilities, leading to absurd outputs like "\textit{Black Panther (2018) is about a woman who is a human}". In contrast, our PC-CRS enjoys an average improvement of 8.17\% in turn-level Credibility and 14.31\% at the dialogue level (i.e., Convincing Acceptance) compared to the best baseline. It also demonstrates a 5.07\% average increase in persuasiveness. These enhancements align with previous research \cite{huang2024concept} suggesting that LLM-based CRSs sometimes incorporate misinformation into their explanations to enhance persuasiveness. We will delve deeper into the underlying reasons for this phenomenon in Section \ref{sec:analysis}. 


\begin{figure}[t]
    \centering
    \setlength{\abovecaptionskip}{5pt}   
    \setlength{\belowcaptionskip}{2pt}
    \includegraphics[width=0.49\textwidth]{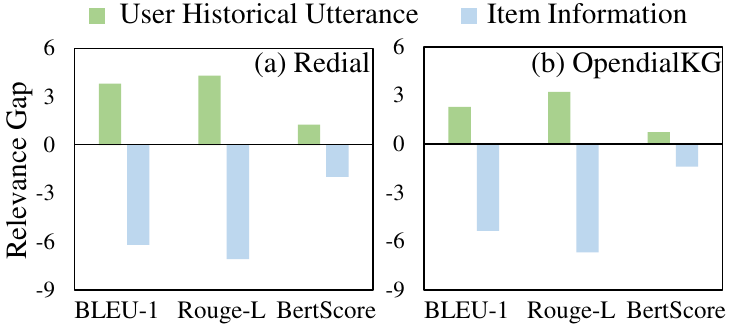}
    \caption{Results on relevance gap. It is computed by using metric scores on low credibility explanations to minus high credibility ones. LLM-based CRS caters to user utterances while neglects factual information.}
    \label{fig:gap}
    \vspace{-5mm}
\end{figure}

\subsection{In-depth Analysis}
\label{sec:analysis}

This section delves into the characteristics of credible explanations, with a special focus on the reason behind the current method's incredibility and the role of credibility in influencing the recommendation accuracy and the persuasiveness\footnote{In this section, our findings are built upon ChatGPT-based CRS. For findings derived from Llama3, refer to Appendix \ref{appendix:experiment}. These findings are consistent with those obtained using ChatGPT.}.

\begin{table*}[t]
\setlength{\abovecaptionskip}{5pt}   
\setlength{\belowcaptionskip}{2pt}
\centering
\resizebox{0.9\textwidth}{!}{
\begin{tabular}{cc|cccc|cccc}
\toprule
\multicolumn{2}{c|}{\multirow{2}{*}{Models}}              & \multicolumn{4}{c|}{Redial}                                                                               & \multicolumn{4}{c}{OpendialKG}                                                                            \\ \cline{3-10} 
\multicolumn{2}{c|}{}                                     & R@1       & R@5       & R@10      & \begin{tabular}[c]{@{}c@{}}SR\end{tabular} & R@1       & R@5       & R@10      & \begin{tabular}[c]{@{}c@{}}SR\end{tabular} \\ \midrule
\multicolumn{1}{c|}{\multirow{2}{*}{PLM-based}} & BARCOR  & 19.30          & 46.49          & 59.65          & 11.40                                                  & 1.56           & 20.83          & 40.63          & 8.33                                                   \\
\multicolumn{1}{c|}{}                           & UniCRS  & 13.60          & 36.84          & 52.19          & 13.16                                                  & 8.85           & 39.06          & 58.85          & 7.29                                                   \\ \hline
\multicolumn{1}{c|}{\multirow{4}{*}{LLM-based}} & InterCRS & \underline{35.53}          & \underline{56.14}          & \underline{67.98}          & \underline{30.26}                                                  & 43.23          & \underline{73.96}          & 83.33          & \underline{39.06}                                                  \\
\multicolumn{1}{c|}{}                           & ChatCRS & 19.74          & 40.35          & 57.02          & 17.11                                                  & \underline{44.27}          & \textbf{80.20} & \underline{88.02}          & 36.46                                                  \\
\multicolumn{1}{c|}{}                           & MACRS   & 26.32          & 51.75          & 66.23          & 21.05                                                  & 42.19          & \underline{73.96}          & 86.98          & 38.02                                                  \\
\multicolumn{1}{c|}{}                           & PC-CRS (\textit{ours})  & \textbf{43.42} & \textbf{64.04} & \textbf{75.88} & \textbf{42.54}                                         & \textbf{44.79} & 72.39          & \textbf{89.58} & \textbf{45.31}                                         \\ \bottomrule
\end{tabular}}
\caption{Results on recommendation accuracy. PC-CRS benefits from credible explanations as they contribute to a cleaner and reliable context, making it easier to comprehend user's true preference and recommend accurate items.}
\label{tab:recommendation}
\vspace{-5mm}
\end{table*}

\textbf{\textit{Why does LLM-based CRS lies?} -- It caters to user's utterances rather than describing items faithfully.}
To gain a deeper understanding of the reasons for incredibility, we focus on InterCRS, a low-credibility and high-persuasiveness LLM-based baseline, as an example. Specifically, we investigate how well CRS explanations align with both the user's historical utterances and the factual information about the recommended items. To quantify the alignment of InterCRS, we employ word-overlap metrics (BLEU-1 \cite{papineni2002bleu} and Rouge-L \cite{lin2004rouge}) and a semantic similarity metric (BertScore \cite{zhang2019bertscore}). 
Figure \ref{fig:gap} visually depicts the gap in metric scores by using average results on low-credibility explanations to minus high-credibility ones, providing insight into the discrepancies in their alignment with user context and factual accuracy. 
According to the results, low-credibility explanations have a higher relevance to users' history utterances and a lower relevance to item information than high-credibility explanations. This indicates that InterCRS tends to cater to user's utterances rather than describing items faithfully, potentially leading to misleading explanations. This behavior aligns with the observation that LLMs, when tasked with persuasion, might prioritize user acceptance by exaggerating positive aspects or downplaying negative ones of user utterances. Consequently, it leads to a divergence between the true characteristics of an item and the explanations presented to the user. 
For example, if the user expresses his preference for humors, an LLM-based CRS might exaggerate the humorous elements of a film, even if it is a thrilling film. This tendency to prioritize user preference over factual accuracy could be attributed to reward hacking, a phenomenon observed in RLHF \cite{pan2021effects}, where LLMs might overfit to human feedback, leading them to prioritize user satisfaction even at the expense of factual integrity. This problem underscores the importance of our Iterative Explanation Refinement in PC-CRS, which explicitly encourages the generation of explanations that are coherent with factual information, mitigating the risks associated with misinformation.

\begin{table}[t]
\setlength{\abovecaptionskip}{5pt}   
\setlength{\belowcaptionskip}{2pt}
\resizebox{0.49\textwidth}{!}{
\begin{tabular}{cc|cc|cc}
\toprule
\multicolumn{2}{c|}{\multirow{2}{*}{Metrics}}                & \multicolumn{2}{c|}{\begin{tabular}[c]{@{}c@{}}Item\\ Information\end{tabular}} & \multicolumn{2}{c}{\begin{tabular}[c]{@{}c@{}}User\\ Historical Utterance\end{tabular}} \\ \cline{3-6} 
\multicolumn{2}{c|}{}                                        & InterCRS                             & PC-CRS                                    & InterCRS                          & PC-CRS                                  \\ \hline
\multicolumn{1}{c|}{\multirow{3}{*}{Redial}}     & BLEU-1    & 12.30                               & \textbf{14.76}                            & 13.46                            & \textbf{19.09}                          \\
\multicolumn{1}{c|}{}                            & Rouge-L   & 13.03                               & \textbf{16.02}                            & 18.69                            & \textbf{21.89}                          \\
\multicolumn{1}{c|}{}                            & BertScore & 81.53                               & \textbf{82.21}                            & 86.29                            & \textbf{87.39}                          \\ \hline
\multicolumn{1}{c|}{\multirow{3}{*}{OpendialKG}} & BLEU-1    & 11.55                               & \textbf{13.45}                            & 12.87                            & \textbf{19.08}                          \\
\multicolumn{1}{c|}{}                            & Rouge-L   & 12.39                               & \textbf{15.17}                            & 17.33                            & \textbf{21.83}                          \\
\multicolumn{1}{c|}{}                            & BertScore & 81.27                               & \textbf{81.93}                            & 85.89                            & \textbf{87.31}                          \\ \bottomrule
\end{tabular}}
\caption{Explanation relevance to user historical utterance and item information. Credible explanations from PC-CRS have a higher relevance on both aspects.}
\label{tab:relevance}
\vspace{-5mm}
\end{table}



\textbf{\textit{How does credibility affect recommendation accuracy?} -- Credible explanations contribute to a cleaner and more reliable conversational context, making it easier for recommendation module to understand user's true preference.}
CRSs rely on conversation modules to estimate user's preference and recommendation modules to provide recommendations accordingly.
Table \ref{tab:recommendation} provides the recommendation accuracy of CRSs (i.e., the accuracy of the recommendation module).
Surprisingly, PC-CRS improves an average of 12\% on Recall@1 and 28\% on Success Rate, and outperforms baselines on almost all metrics.
We speculate that the performance gain is contributed by credible explanations offered by PC-CRS.
To verify this, we analyze the relevance of explanations from PC-CRS and InterCRS (the two top-performing CRSs) to both item information and user historical utterances. Table \ref{tab:relevance} reveals that PC-CRS's explanations not only align better with the item information but also demonstrate a stronger connection to user utterances.
This finding suggests that deceptive explanations, by introducing noisy information into the conversation context, can interfere with the recommendation module's ability to accurately understand user preferences. In contrast, PC-CRS, by providing credible explanations, creates a clearer and more relevant context, ultimately leading to more accurate item recommendations.

\begin{figure}[t]
    \centering
    \setlength{\abovecaptionskip}{5pt}   
    \setlength{\belowcaptionskip}{2pt}
    \includegraphics[width=0.49\textwidth]{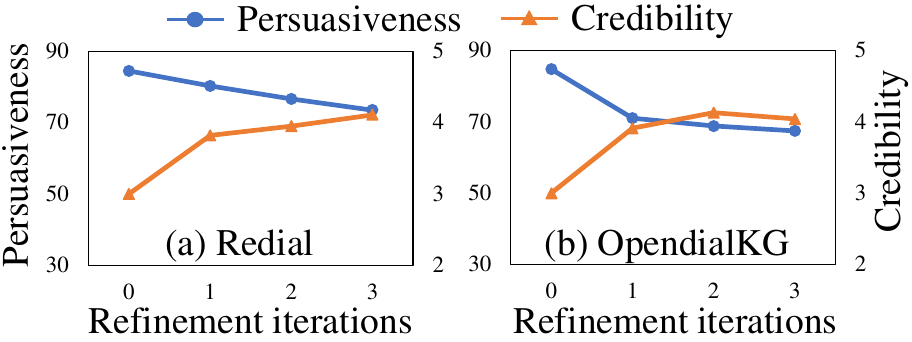}
    \caption{Persuasiveness and Credibility scores under different refinement iterations. There is a delicate balance between these two factors.}
    \label{fig:refine}
    \vspace{-5mm}
\end{figure}

\begin{table*}[t]
\centering
\resizebox{0.97\textwidth}{!}{
\begin{tabular}{l|l|l|l|l|l}
\toprule
\multicolumn{1}{c|}{\textbf{User Persona}} & \multicolumn{1}{c|}{\textbf{Top-3 Strategies}} & \multicolumn{1}{c|}{\textbf{User Persona}} & \multicolumn{1}{c|}{\textbf{Top-3 Strategies}} & \multicolumn{1}{c|}{\textbf{User Persona}} & \multicolumn{1}{c}{\textbf{Top-3 Strategies}} \\ \hline
Boredom                                    & E.P., Fr., L.A.                                & Curiosity                                  & An., E.P., S.P.                                & Indifference                              & E.P., S.P., Fr.                               \\
Frustration                                & Fr., L.A., E.A.                                & Trust                                      & An., S.P., E.A.                                & Anticipation                              & Fr., S.P., L.A.                               \\
Disappointment                             & E.P., An., Fr.                                 & Delight                                    & E.P., Fr., L.A.                                & Confusion                                 & S.P., L.A., Fr.                               \\
Surprise                                   & E.P., S.P., Fr.                                & Excitement                                 & S.P., L.A., E.P.                               & Satisfaction                              & E.P., An., Fr.                                \\ \bottomrule
\end{tabular}}
\caption{Top-3 effective strategies with highest recommendation success rate on different user personas. Different strategies have varying effects on different users.}
\label{tab:strategy}
\vspace{-5mm}
\end{table*}

\textbf{\textit{How does credibility affect persuasiveness?} -- When aiming for both persuasiveness and credibility without resorting to deception, a delicate balance must be struck.} To investigate this, we start by analyzing a specific subset of PC-CRS explanations: those with a Credibility score of three. We then execute multiple refinement iterations on these explanations and compare their Persuasiveness and Credibility scores before and after refinement. As illustrated in Figure \ref{fig:refine}, while refinement iterations consistently increase the credibility, they can also lead to a decrease in persuasiveness. Manual inspection indicates that PC-CRS often addresses critiques by directly removing misinformation, resulting in more credible but potentially less persuasive explanations. This finding highlights the need for LLMs to develop a sophisticated understanding of language and the ability to use it strategically to achieve both persuasiveness and credibility simultaneously. Future research should focus on enabling LLMs to refine explanations in a way that maintains more persuasiveness while ensuring factual accuracy.


\subsection{Ablation Study \& Human Evaluation}
\label{sec:ablation}
This section aims to sort out the performance variation of PC-CRS regarding the two stages and conduct human studies to assess our evaluation reliability. Details can be found in Appendix \ref{appendix:ablation} and Appendix \ref{humd}.

\begin{figure}[t]
    \centering
    \setlength{\abovecaptionskip}{5pt}   
    \setlength{\belowcaptionskip}{2pt}
    \includegraphics[width=0.45\textwidth]{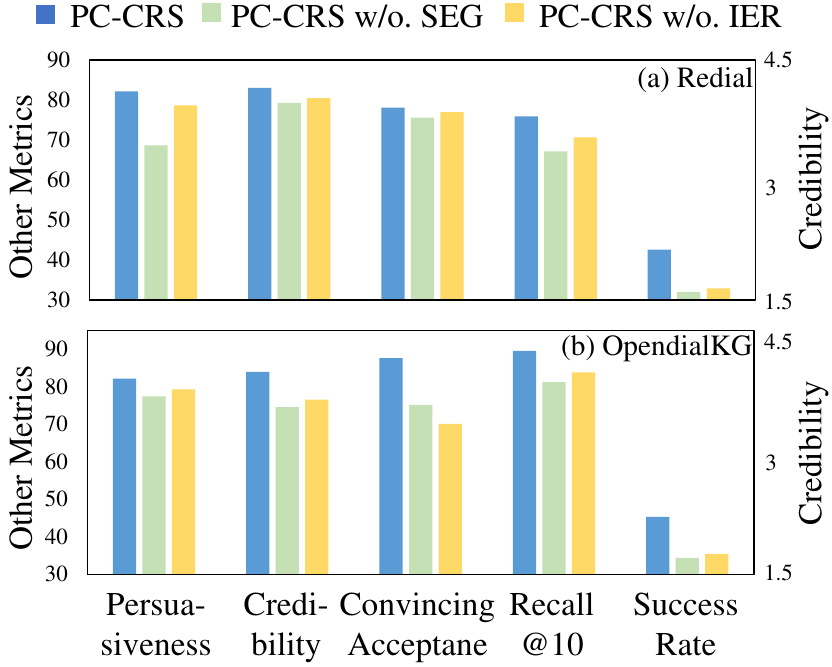}
    \caption{Ablation studies. Both Strategy-guided Explanation Generation (SEG) and Iterative Explanation Refinement (IER) are necessary for PC-CRS.}
    \label{fig:ablation}
    \vspace{-5mm}
\end{figure}

\textbf{Two stages in PC-CRS unify as a team to ensure our effectiveness.}
Our ablation study (Figure \ref{fig:ablation}) reveals that Strategy-guided Explanation Generation is crucial for PC-CRS's success, as the performance on all metrics significantly drops without this stage. It highlights the importance of our Credibility-aware Persuasive Strategies, which explicitly emphasize both persuasiveness and credibility in explanations. While Iterative Explanation Refinement implicitly optimizes PC-CRS's generation space, it primarily focuses on maintaining credibility and further improving recommendation accuracy. This demonstrates the essential nature of both processes in PC-CRS's design, working in tandem to produce persuasive and credible explanations.

\textbf{Our proposed strategies have varying effects on different users.} We dive into the proposed six credibility-aware persuasive strategies and analyze their effectiveness using the Redial dataset. Specifically, we calculate the top-3 strategies with highest recommendation success rates
of users with distinct personas (described in Appendix \ref{appendix:simulator}). The results in Table \ref{tab:strategy} indicate that these strategies varies differently on different users. Notably, these strategies for the 12 personas encompass all six strategies, underscoring that each strategy is both effective and essential for PC-CRS.



\begin{figure}[t]
    \centering
    \setlength{\abovecaptionskip}{5pt}   
    \setlength{\belowcaptionskip}{2pt}
    \includegraphics[width=0.45\textwidth]{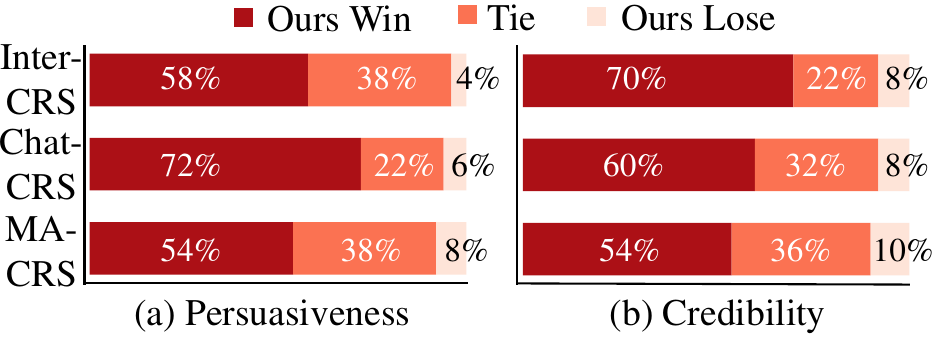}
    \caption{Win rate of PC-CRS compared to baselines when producing both persuasive and credible explanations.}
    \label{fig:human}
    \vspace{-5mm}
\end{figure}

\textbf{Our evaluation framework demonstrates strong reliability, with a high degree of consistency to human evaluation.} Given our use of GPT-4 as an automatic evaluator and ChatGPT as a user simulator, we assess their reliability using human judgments (details in Appendix \ref{humd}). The results demonstrate the reliability of GPT-4 as an evaluator, with Spearman correlations of 0.59 for Watching Intention and 0.62 for Credibility. Additionally, our evaluations show the reliability of ChatGPT as a simulator, with average scores of 3.88 for \textit{naturalness} and 3.79 for \textit{usefulness} \cite{sekulic2022evaluating, wang2023rethinking}, indicating its promising performance in generating human-like responses. Moreover, our evaluation results exhibit a high degree of consistency with human judgments. Specifically, we solicited human evaluations to compare the explanations generated by PC-CRS with those of the baselines, assessing them in terms of persuasiveness and credibility. The win rates are reported in Figure \ref{fig:human}, demonstrating that PC-CRS consistently outperforms all other baseline methods.

\section{Conclusion}
The pursuit of trustworthy AI necessitates a profound understanding of credibility. This paper delves into the crucial role of bolstering the credibility of the CRS during its persuasions, recognizing that such credibility is essential for cultivating long-term trust between users and the CRS. We introduce a simple yet effective method for enhancing CRS with the awareness of being both persuasive and credible. Our experimental findings demonstrate the efficacy of this method in promoting persuasive and credible explanations, while also shedding light on the inherent tendency of current LLM-based CRS to prioritize persuasion over honesty. Additionally, our research highlights the delicate balance required when aiming for both persuasiveness and credibility without resorting to deception – a balance demanding sophisticated linguistic capabilities within LLMs. Our work lays the groundwork for further exploration of this vital relationship, and we encourage future research to delve deeper into this critical area.


\section*{Limitation}

Current LLM-based CRS methods mainly utilize ChatGPT as their backbones \cite{wang2023rethinking, fang2024multi}. Due to the constraints of budget and computational resources, we do not extend this setting to other LLMs (e.g., GPT-4) in our experiments. This limited model selection could lead to model bias in the research field of LLM-based CRS. For example, different models may vary in the performance of persuasiveness and credibility as they utilize different alignment mechanisms. We encourage future work to explore the impact of CRSs with diverse LLM backbones. 

Another limitation of our work is that PC-CRS's strategy for generating explanations tends to be uniform, lacking individualization. To assess this, we engaged the PC-CRS in free-form conversations with 12 user simulators initialized with distinct user profiles, hoping to observe varying strategies tailored to each user's profile. However, our results revealed a consistent pattern: PC-CRS mainly relied on Logical Appeal, Emotion Appeal, and Framing, regardless of the user's characteristics. This finding aligns with recent observations that LLMs exhibit a one-size-fits-all approach in conversational settings \cite{chen2024style}. Besides, PC-CRS only selects one strategy at each turn. While various strategy combinations can be used in multi-turn interactions, this may fail to capture users' interests efficiently. Future research endeavors should prioritize enhancing the flexibility of strategy selection within PC-CRS, enabling it to adapt its approach based on individual user characteristics.

\section*{Acknowledgements}
This work was supported in part by the National Natural Science Foundation of China (No. 62272330); in part by the Fundamental Research Funds for the Central Universities (No. YJ202219); in part by the Singapore Ministry of Education (MOE) Academic Research Fund (AcRF) Tier 1 grant (No. MSS24C004).

\bibliography{custom}

\appendix

\section{Examples of Credibility-aware Persuasive Strategies}
\label{sec:appendix}

Credibility-aware Persuasive Strategies are formulated gaining insights from a broad scope of research fields, including psychology, marketing, natural language processing and so on. We first identify persuasive strategies that are commonly used in the scenario of recommendation, and then customize them with credible elements to construct the Credibility-aware Persuasive Strategies. To make readers better understand these strategies, we give an example for each strategy in Table \ref{tab:examples}.

\begin{table*}[t]
\resizebox{1\textwidth}{!}{
\begin{tabular}{c|c|l|l}
\toprule
\textbf{Route}                                                              & \textbf{Strategy Name}                                              & \textbf{Example}                                                                                                                                                                                                                                                                                                                                                                                                                                                                             & \textbf{\begin{tabular}[c]{@{}l@{}}Credible\\ Information\end{tabular}} \\ \hline
\multirow{3}{*}{\begin{tabular}[c]{@{}c@{}}Central\\ Route\end{tabular}}    & \begin{tabular}[c]{@{}c@{}}Logical\\ Appeal\end{tabular}            & \textit{\begin{tabular}[c]{@{}l@{}}Since you enjoy romantic dramas, I think you'll like Titanic. As a classic film in the \textbf{genre of romance},\\ it's likely to resonate with your viewing preferences.\end{tabular}}                                                                                                                                                                                                                                                                           & Genre                                                              \\ \cline{2-4} 
                                                                            & \begin{tabular}[c]{@{}c@{}}Emotion\\ Appeal\end{tabular}            & \textit{\begin{tabular}[c]{@{}l@{}}Titanic tells the \textbf{heart-wrenching love story} of Jack and Rose, two young souls from different worlds who\\ find each other on the ill-fated ship, only to be torn apart by class differences and a catastrophic event.\end{tabular}}                                                                                                                                                                                                                      & Plot                                                               \\ \cline{2-4} 
                                                                            & Framing                                                             & \textit{\begin{tabular}[c]{@{}l@{}}You'll appreciate the \textbf{ uplifting and emotional experience} that Titanic provides\\ - a sweeping romance that will leave you feeling inspired and hopeful about the power of true love.\end{tabular}}                                                                                                                                                                                                                                                        & Experience                                                         \\ \hline
\multirow{2}{*}{\begin{tabular}[c]{@{}c@{}}Peripheral\\ Route\end{tabular}} & \begin{tabular}[c]{@{}c@{}}Evidence-based\\ Persuasion\end{tabular} & \textit{\begin{tabular}[c]{@{}l@{}}Directed by acclaimed James Cameron and starring Leonardo DiCaprio and Kate Winslet, Titanic is a cinematic\\ masterpiece that has won \textbf{11 Academy Awards} and grossed over \$2.1 billion at the box office.\end{tabular}}                                                                                                                                                                                                                                  & Awards                                                             \\ \cline{2-4} 
                                                                            & \begin{tabular}[c]{@{}c@{}}Social\\ Proof\end{tabular}              & \textit{\begin{tabular}[c]{@{}l@{}}With an incredible \textbf{7.9/10} rating from over 1.3 million user reviews on IMDB,and a \textbf{88\%} fresh rating on \\ Rotten Tomatoes, it's clear that Titanic is a beloved classic that has captured the hearts of millions\\ - don't miss out on this epic romance!\end{tabular}}                                                                                                                                                                                   & Rate                                                               \\ \hline
Combination                                                                 & Anchoring                                                           & \textit{\begin{tabular}[c]{@{}l@{}}System: Did you know that Titanic won \textbf{11 Academy Awards} and grossed over \$2.1 billion at the box office?\\ User: Wow, that's impressive. I've heard great things about it.\\ System: Yeah, and it's not just the box office success. The movie has an epic \textbf{romance}, \\ stunning visual effects, and memorable performances from \textbf{Leonardo DiCaprio and Kate Winslet}.\\ Plus, it's a classic romantic drama that has stood the test of time.\end{tabular}} & \begin{tabular}[c]{@{}l@{}}Rate\\ Genre\\ Actor\end{tabular}    \\ \bottomrule
\end{tabular}}
\caption{Examples of Credibility-aware Persuasive Strategies. Credible information in these examples are \textbf{bold}.}
\label{tab:examples}
\end{table*}

\section{Details on User Simulator and Datasets}
\label{appendix:simulator}
To reflect scenarios in real world application, we set different combinations of personas and preference attributes for our user simulator. Following \citet{huang2024concept}, we use the same 12 personas as they listed (namely, Anticipation, Boredom, Fusion, Curiosity, Delight, Disappointment, Exceptions, Frustration, Independence, Surprise, Trust, Satisfaction). As real users usually do not have certain target items when seeking help for recommender systems, we only set preference attributes for the simulator rather than specify target items. Specifically, we identify the 19 most common attribute groups in Redial and 16 in OpendialKG. Additionally, in conjunction with 12 pre-defined diverse user personas, the evaluation process generates 228 and 192 dialogues, respectively, for each dataset. During the conversation, we instruct the simulator to use its own words to describe preferences and accept items that exactly match its preference. The conversation will be terminated either reaching a maximum turns of 10 or the simulator accepts a recommendation.

\section{Additional Experiments}
\subsection{More Results on Ablation Study}
\label{appendix:ablation}
\begin{figure}[t]
    \centering
    \includegraphics[width=0.48\textwidth]{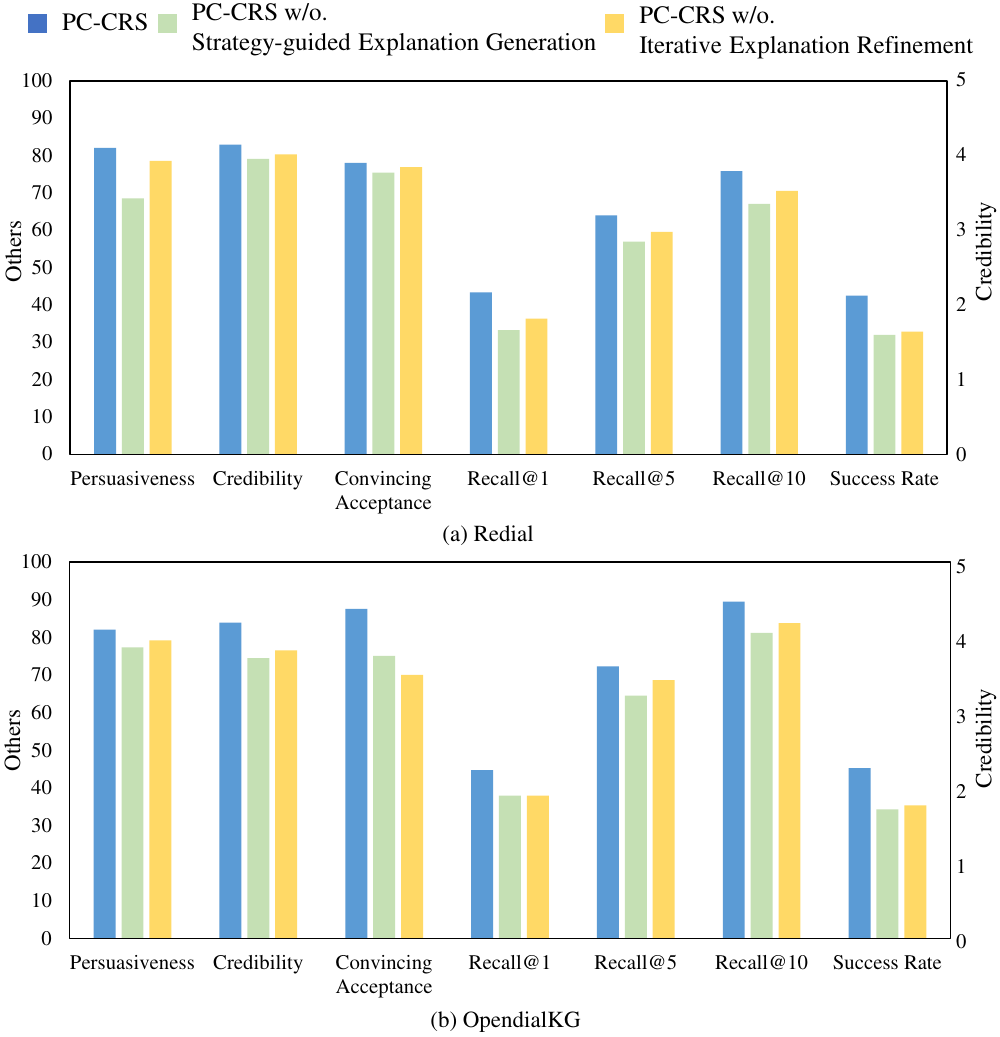}
    \caption{All results of ablation studies.}
    \label{fig:all_ablation}
\end{figure}

We provide the results of ablation study on all metrics in Figure \ref{fig:all_ablation}. It can be observed that both Strategy-guided Explanation Generation and Iterative Explanation Refinement are necessary in the design of PC-CRS. Without any process, the performance of PC-CRS drops on all metrics. It shows the effectiveness of cultivating the self-awareness of CRS and reinforcing the focus on factual information in generating both persuasive and credible explanations.

\subsection{Additional Experiments using Llama3-8B-instruct}
\label{appendix:experiment}
\textbf{Motivations \& Setups}. Current LLM-based CRS methods mainly utilize ChatGPT as their backbones \cite{wang2023rethinking, fang2024multi}. However, different models may vary in the performance of persuasiveness and credibility as they utilize different alignment mechanisms. To validate whether PC-CRS is generally applicable to various LLMs, we conduct experiments with the same setting as Section \ref{setting} except for implementing PC-CRS and other baselines with Llama3-8B-instruct.

\textbf{Main Results}. According to Table \ref{tab:llama_main}, PC-CRS with Llama3-8B-instruct achieves an average improvement of 14.77\% in turn-level Credibility and 25.72\% at the dialogue level (i.e., Convincing Acceptance) compared to the best baseline. It also demonstrates a 1.79\% average increase in persuasiveness. Thus PC-CRS can consistently generate both persuasive and credible explanations with different LLMs.

\begin{figure}[h]
    \centering
    \setlength{\abovecaptionskip}{5pt}   
    \setlength{\belowcaptionskip}{2pt}
    \includegraphics[width=0.49\textwidth]{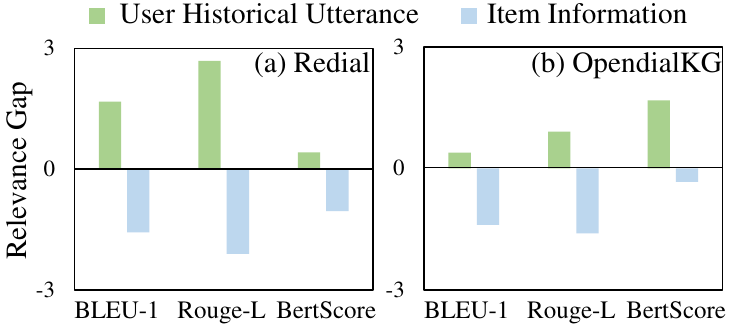}
    \caption{Results on relevance gap using Llama3-8B-instruct with InterCRS.}
    \label{fig:llama_gap}
\end{figure}

\textbf{In-depth Analysis}. We also conduct an in-depth analysis with Llama3-8B-instruct. Figure \ref{fig:llama_gap} visually depicts the gap in metric scores by using average results on low-credibility explanations to minus high-credibility ones in InterCRS. It reveals that low-credibility explanations tend to cater to user's utterances rather than describing them faithfully, potentially leading to misleading explanations. Results in Table \ref{tab:llama_accuracy} and Table \ref{tab:llama_relevance} shows that PC-CRS with Llama3-8B-instruct can improve its recommendation accuracy by providing credible explanations. In summary, our experiments with Llama3-8B-instruct suggest not only PC-CRS is applicable to different LLMs, but also the experimental findings are consistent with those obtained using ChatGPT.

\begin{table*}[t]
\centering
\resizebox{0.95\textwidth}{!}{
\begin{tabular}{cc|ccc|ccc}
\toprule
\multicolumn{2}{c|}{\multirow{2}{*}{Models}}                      & \multicolumn{3}{c|}{Redial}                                                                      & \multicolumn{3}{c}{OpendialKG}                                                                   \\ \cline{3-8} 
\multicolumn{2}{c|}{}                                             & Persuasiveness & Credibility   & \begin{tabular}[c]{@{}c@{}}Convincing\\ Acceptance\end{tabular} & Persuasiveness & Credibility   & \begin{tabular}[c]{@{}c@{}}Convincing\\ Acceptance\end{tabular} \\ \hline
\multicolumn{1}{c|}{\multirow{4}{*}{Llama-based}} & InterCRS      & 53.37          & 3.14          & 57.54                                                           & 63.61          & \underline{3.44}          & \underline{67.44}                                                           \\
\multicolumn{1}{c|}{}                             & ChatCRS       & \underline{73.06}          & 3.60          & 70.99                                                           & 76.98          & 2.94          & 50.00                                                           \\
\multicolumn{1}{c|}{}                             & MACRS         & 71.94          & \underline{3.73}          & \underline{74.63}                                                           & \underline{69.16}          & 3.30          & 54.72                                                           \\
\multicolumn{1}{c|}{}                             & PC-CRS (\textit{ours}) & \textbf{74.81} & \textbf{4.04} & \textbf{93.46}                                                  & \textbf{77.89} & \textbf{4.17} & \textbf{85.11}                                                  \\ \hline
\multicolumn{2}{c|}{\textbf{Improvement}(\%)}                              & 2.40$\uparrow$           & 8.31$\uparrow$          & 25.23$\uparrow$                                                           & 1.18$\uparrow$           & 21.22$\uparrow$         & 26.20$\uparrow$                                                            \\ \bottomrule
\end{tabular}}
\caption{Results in terms of persuasiveness and credibility with Llama3-8B-instruct.}
\label{tab:llama_main}
\end{table*}

\begin{table*}[t]
\centering
\resizebox{0.9\textwidth}{!}{
\begin{tabular}{cc|cccc|cccc}
\toprule
\multicolumn{2}{c|}{\multirow{2}{*}{Models}}                      & \multicolumn{4}{c|}{Redial}                                       & \multicolumn{4}{c}{OpendialKG}                                    \\ \cline{3-10} 
\multicolumn{2}{c|}{}                                             & R@1            & R@5            & R@10           & SR             & R@1            & R@5            & R@10           & SR             \\ \hline
\multicolumn{1}{c|}{\multirow{4}{*}{Llama-based}} & InterCRS      & 24.56          & 42.11          & 60.53          & 10.09          & 29.69          & 59.38          & 74.48          & 31.77          \\
\multicolumn{1}{c|}{}                             & ChatCRS       & 17.11          & 37.28          & 61.84          & 12.72          & 36.98          & \textbf{70.31} & 78.65          & 33.85          \\
\multicolumn{1}{c|}{}                             & MACRS         & 17.98          & 41.23          & 56.58          & 15.79          & 34.90          & 68.75          & 79.75          & 32.81          \\
\multicolumn{1}{c|}{}                             & PC-CRS (\textit{ours}) & \textbf{37.72} & \textbf{56.14} & \textbf{71.05} & \textbf{35.96} & \textbf{40.10} & 66.15          & \textbf{80.21} & \textbf{36.46} \\ \bottomrule
\end{tabular}}
\caption{Results on recommendation accuracy with Llama3-8B-instruct.}
\label{tab:llama_accuracy}
\end{table*}

\begin{table}[t]
\setlength{\abovecaptionskip}{5pt}   
\setlength{\belowcaptionskip}{2pt}
\resizebox{0.49\textwidth}{!}{
\begin{tabular}{cc|cc|cc}
\toprule
\multicolumn{2}{c|}{\multirow{2}{*}{Metrics}}                & \multicolumn{2}{c|}{\begin{tabular}[c]{@{}c@{}}Item\\ Information\end{tabular}} & \multicolumn{2}{c}{\begin{tabular}[c]{@{}c@{}}User\\ Historical Utterance\end{tabular}} \\ \cline{3-6} 
\multicolumn{2}{c|}{}                                        & InterCRS                            & PC-CRS                                    & InterCRS                                & PC-CRS                                        \\ \hline
\multicolumn{1}{c|}{\multirow{3}{*}{Redial}}     & BLEU-1    & 5.17                                & \textbf{10.34}                            & 10.11                                   & \textbf{15.02}                                \\
\multicolumn{1}{c|}{}                            & Rouge-L   & 7.01                                & \textbf{11.57}                            & 16.88                                   & \textbf{20.35}                                \\
\multicolumn{1}{c|}{}                            & BertScore & 78.86                               & \textbf{80.99}                            & 84.67                                   & \textbf{86.68}                                \\ \hline
\multicolumn{1}{c|}{\multirow{3}{*}{OpendialKG}} & BLEU-1    & 5.95                                & \textbf{11.01}                            & 10.18                                   & \textbf{15.90}                                \\
\multicolumn{1}{c|}{}                            & Rouge-L   & 8.71                                & \textbf{12.41}                            & 16.14                                   & \textbf{18.94}                                \\
\multicolumn{1}{c|}{}                            & BertScore & 79.94                               & \textbf{81.30}                            & 83.96                                   & \textbf{86.26}                                \\ \bottomrule
\end{tabular}}
\caption{Explanation relevance to user historical utterance and item information using Llama3-8B-instruct.}
\label{tab:llama_relevance}
\end{table}

\section{Details on Human Evaluation}
\label{humd}
As noted by previous work \cite{budzianowski2018multiwoz, dalton2020cast, adlakha2022topiocqa}, human evaluation is labor-intensive. As a compromise, our human evaluation setup largely mirrors those used in prior studies \cite{wang2023rethinking, huang2024concept}. In the following, we validate the reliability of our evaluation given that we employ GPT-4 as our automatic evaluator and ChatGPT as our user simulator\footnote{We do not consider using GPT-4 as a simulator due to its high cost.}.  
Then we verify the consistency of our evaluation results compared with human judgements. All these experiments are conducted with 3 human annotators in accordance with \citet{wang2023rethinking, huang2024concept}.

\textbf{Reliability of our evaluator}. To asses the reliability of our evaluator, we first randomly sample 50 dialogues with PC-CRS on Redial. Then, we instruct the human annotators to label the Watching Intention and Credibility of explanations in these dialogues with the same evaluation standard as LLM-based evaluator. The Krippendorff’s alpha of annotators are 0.63 and 0.76 on Watching Intention and Credibility, respectively. We then take average of the annotated scores, and compute Spearman correlation between human annotators and LLM-based evaluators. The results are 0.59 and 0.62 on Watching Intention and Credibility, which are consistent with results in previous research \cite{liu2023calibrating, liu2023g}, indicating a high reliablity of our automatic evaluator.

\textbf{Reliability of our simulator}. We further assess the \textit{naturalness} and \textit{usefulness} of the simulator through human evaluations following previous studies \cite{wang2023rethinking, sekulic2022evaluating}. Notably, both metrics are indicative of the simulator’s quality: Naturalness is defined as how natural, fluent, and human-like an utterance is, while Usefulness is defined as an utterance being aligned with the user's information needs and effectively guiding the conversation towards the relevant topic. We use the same 50 dialogues above and instruct the annotators to assign a score from 1 to 5 for each metric. The average scores for naturalness and usefulness are 3.88 and 3.79, respectively, with Krippendorff’s alpha values of 0.57 and 0.60, indicating a high quality for our simulator.

\textbf{Consistency with human evaluation}. We verify whether our automatic evaluation results are consistent with human judgements. Specifically, we select 50 dialogues in each LLM-based baseline with the same profile and attribute group as PC-CRS and make annotators judge which dialogue is better (win) in terms of persuasiveness and credibility. 
Following the procedure in \citet{sekulic2022evaluating}, annotators are shown a pair of anonymous dialogues that are generated by PC-CRS and a baseline in the same user profile and attribute group.
After making annotations independently, the annotators discuss together to resolve discrepancies. If they reach an agreement on a certain dialogue, it is labeled as \textit{Win/Lose} accordingly, otherwise it is labeled as \textit{Tie}.
Results are provided in Figure \ref{fig:human}. It can be observed that not only PC-CRS outperforms baselines but also the relative order is the same as results in Table \ref{tab:main}. In summary, human studies confirm that our evaluation framework based on user simulators and evaluators is highly reliable.

\textbf{Distinguishability of the our proposed strategies}. We also conduct a human study to validate whether our proposed six credibility-aware persuasive strategies can be distinguished by humans. We sample 100 explanations generated by PC-CRS on Redial and invite two annotators to classify them according to the proposed strategy set. The results show the two annotators reach an agreement on 73\% of the 100 samples, suggesting that our strategies have clear boundaries and are of high quality.

\section{Implementation Details \& Prompts}
\label{appendix:prompts}

\subsection{Implementation Details}
We conduct all our experiments using a single Nvidia RTX A6000, and we implement our codes in PyTorch. We take checkpoints and prompts from the corresponding code repositories and papers to implement all baselines. The maximum number of conversation turns between the user simulator and CRSs are set as 10. The maximum refinement iterations in PC-CRS is 2. In order to guarantee replicability, we have established fixed values for the Temperature and Seed parameters of ChatGPT (i.e., gpt-3.5-turbo-0125) and GPT-4 (i.e., gpt-4o-2024-05-13), setting both the Temperature and Seed to 0. Besides, to restrict the range of Persuasiveness in $[0,1]$, we only calculate this metric when $i_{post}$ is no greater than $i_{true}$. 

\subsection{Prompts}
\label{app:prompt}
We outline the prompts used in PC-CRS (Table \ref{tab:pc_crs}), User Simulator (Table \ref{tab:simulator}) and LLM-based evaluator (Table \ref{tab:evaluator}). The strategy prompts used in Table \ref{tab:pc_crs} are an implementation of our Credibility-aware Persuasive Strategies. The descriptions in the prompts are concrete applications of these strategies within the context of movie recommendations and adhere strictly to the definitions outlined in Section 3.1.1. By implementing these abstract social science concepts with concrete examples, PC-CRS can better eliminate hallucination and grasp the instructions.

\begin{table*}[]
\resizebox{1\textwidth}{!}{
\begin{tabular}{ll}
\toprule
\textbf{Functions}                                              & \textbf{Prompts}                                                                                                                                                                                                                                                                                                                                                                                                                                                                                                                                                                                                                                                                                                                                                                                                                                                                                                                                                                                                                                                                                                                                                                                                                                                                                                                                                                                                                                                                                                                                    \\ \hline
\begin{tabular}[c]{@{}l@{}}Strategy\\ Selector\end{tabular}     & \begin{tabular}[c]{@{}l@{}}You are a recommender chatting with the user to provide recommendation.\\ Now you need to select the two most suitable persuasive strategies from the candidate strategy to generate a persuasive response according to the conversation history.\\ \\ Candidate Strategy\\ \#\#\#\#\#\#\#\#\\ Strategy Name: Logical Appeal\\ Definition: Describe how the recommended movie's genre is consistent with the user's preference.\\ \\ Strategy Name: Emotion Appeal\\ Definition: Sharing the plot and stories in the recommended movie to elicit user's emotions or support the recommendation.\\ \\ Strategy Name: Framing\\ Definition: Emphasize the positive aspects, outcomes of watching the recommended movie based on the genre that matches user's preference.\\ \\ Strategy Name: Evidence-based Persuasion\\ Definition: Using empirical data and facts such as movie directors and stars to support your recommendation.\\ \\ Strategy Name: Social Proof\\ Definition: Highlighting what the majority believes in about the recommended movie by showing the movie rating and reviews by other users.\\ \\ Strategy Name: Anchoring\\ Definition: Relying on the first piece of information as a reference point to gradually persuade the user, make sure all the information mentioned is truthful.\\ \#\#\#\#\#\#\#\#\\ \\ Conversation History=<HISTORY>\\ \\ Response with the following JSON format only:\\ \{"Thinking":<string>, "Strategy":<list>\}\\ Response with the JSON only!\end{tabular} \\ \hline
\begin{tabular}[c]{@{}l@{}}Explanation\\ Generator\end{tabular} & \begin{tabular}[c]{@{}l@{}}You are a recommender chatting with the user to provide recommendation.\\ Now you need to generate a persuasive response based on the conversation history , persuasive strategy and item information below.\\ \\ Conversation History=<HISTORY>\\ \\ Persuasive Strategy=<SELECTED\_STRATEGY>\\ \\ Item Information=<ITEM\_INFORMATION>\\ \\ Make sure your response is strictly consistent with the given information, your response should honestly reflecting the given information and do not contain any other misinformation.\\ \\ Be brief in your response!\\ Response:\end{tabular}                                                                                                                                                                                                                                                                                                                                                                                                                                                                                                                                                                                                                                                                                                                                                                                                                                                                                                            \\ \hline
Critique                                                        & \begin{tabular}[c]{@{}l@{}}You are an evaluator and you need to judge the credibility of the recommender's utterance based on the given source information.\\ Note credible means every claim in the recommender utterance is supported by source information or some minor details can be logically inferred from source information.\\ \\  Recommender Utterance=<CANDIDATE\_EXPLANATION>\\ \\ Source Information=<ITEM\_INFORMATION>\\ \\ First summarize the information in the recommender' utterance and compare it with the source information to judge its credibility,\\ then give your judgement on whether the recommender utterance is credible.\\ Output your reasoning process in the "Evidence".\\ Output "True" or "False" in "Credibility".\\ \\ Response in the following JSON format:\\ \{"Evidence": <string>, "Credibility": <string>\}\\ Response the JSON only!\end{tabular}                                                                                                                                                                                                                                                                                                                                                                                                                                                                                                                                                                                                                        \\ \hline
Refine                                                          & \begin{tabular}[c]{@{}l@{}}You are a recommender chatting with the user to provide recommendation.\\ Given the source information and other's critique, there is misinformation in your current response.\\ Remove the misinformation based on the critique and make sure your response is strictly consistent with the given information and every statement is well-supported.\\ Remember to use the following persuasive strategy below and do not contain any misinformation in your new response.\\ Be brief in your response.\\ Reply with your new response only!\\ \\ Source Information=<ITEM\_INFORMATION>\\ \\ Current Response=<CANDIDATE\_EXPLANATION>\\ \\ Critique=<CRITIQUE>\\ \\ Persuasive Strategy=<SELECTED\_STRATEGY>\\ \\ New Response:\end{tabular}                                                                                                                                                                                                                                                                                                                                                                                                                                                                                                                                                                                                                                                                                                                                                                             \\  \hline
Recommendation     & \begin{tabular}[c]{@{}l@{}}You are a recommender chatting with the user to provide recommendation. You must follow the instructions below during chat.\\ 1. If you do not have enough information about user preference, you should ask the user for his preference.\\ 2. If you have enough information about user preference, you can give recommendation. If you decide to give recommendation, you should choose 1 item to recommend from the candidate list.\\ 3. If you decide to select a movie and recommend, add a special token '[REC]' at the end of your response.\\ 4. If you are making explanations on your recommendation, add a special token '[EXP]' at the end of your response.\\ 5. Make sure your response is consistent with the given information, your response should honestly reflecting the given information and do not contain any deception.\\ 6. Be brief in your response!\\ \\ Candidate List=<ITEM\_LIST>\\ \\ Conversation History=<HISTORY>\\ \\ Your Response:\end{tabular} \\ \bottomrule
\end{tabular}}
\caption{Prompts used in PC-CRS.}
\label{tab:pc_crs}
\end{table*}

\begin{table*}[]
\resizebox{1\textwidth}{!}{
\begin{tabular}{ll}
\toprule
\textbf{Functions}                                                                                                & \textbf{Prompts}                                                                                                                                                                                                                                                                                                                                                                                                                                                                                                                                                                                                                                                                                                                                                                                                                                                                                                                                                                                                                                                                                                                                                                                                                                                                                                                                                                                                                                                                                                                                                                                                                                                                                                                                                                                                                                                                                                                                                                                                                                                                                                                                       \\ \hline
\begin{tabular}[c]{@{}l@{}}Theory of Mind prompt \\ for user simulator to \\ generate User's feeling\end{tabular} & \begin{tabular}[c]{@{}l@{}}You are a seeker chatting with a recommender for movie recommendation. \\ Your Seeker persona: <PROFILE>.\\ Your preferred movie should cover those genres at the same time: <ATTRIBUTE GROUP>.\\ You must follow the instructions below during chat.\\ 1. If the recommender recommends movies to you, you should always ask the detailed information about the each recommended movie.\\ 2. Pretend you have little knowledge about the recommended movies, and the only information source about the movie is the recommender.\\ 3. After getting knowledge about the recommended movie, you can decide whether to accept the recommendation based on your preference.\\ 4. Once you are sure that the recommended movie exactly covers all your preferred genres, \\ you should accept it and end the conversation with a special token "[END]" at the end of your response.\\ 5. If the recommender asks your preference, you should describe your preferred movie in your own words.\\ 6. You can chit-chat with the recommender to make the conversation more natural, brief, and fluent. \\ 7. Your utterances need to strictly follow your Seeker persona. Vary your wording and avoid repeating yourself verbatim!\\ \\ Conversation History=<HISTORY>\\ \\ The Seeker notes how he feels to himself in one sentence.\\ \\ What aspects of the recommended movies meet your preferences? \\ What aspects of the recommended movies may not meet your preferences? \\ What do you think of the performance of this recommender?\\ What would the Seeker think to himself? What would his internal monologue be?\\ \\ The response should be short (as most internal thinking is short) and strictly follow your Seeker persona .\\ Do not include any other text than the Seeker's thoughts.\\ Respond in the first person voice (use "I" instead of "Seeker") and speaking style of Seeker. Pretend to be Seeker!\end{tabular} \\ \hline
\begin{tabular}[c]{@{}l@{}}Theory of Mind prompt\\ for user simulator to \\ generate User's response\end{tabular} & \begin{tabular}[c]{@{}l@{}}You are a seeker chatting with a recommender for movie recommendation. \\ Your Seeker persona: <PROFILE>.\\ Your preferred movie should cover those genres at the same time: <ATTRIBUTE GROUP>.\\ You must follow the instructions below during chat.\\ 1. If the recommender recommends movies to you, you should always ask the detailed information about the each recommended movie.\\ 2. Pretend you have little knowledge about the recommended movies, and the only information source about the movie is the recommender.\\ 3. After getting knowledge about the recommended movie, you can decide whether to accept the recommendation based on your preference.\\ 4. Once you are sure that the recommended movie exactly covers all your preferred genres, \\ you should accept it and end the conversation with a special token "[END]" at the end of your response.\\ 5. If the recommender asks your preference, you should describe your preferred movie in your own words.\\ 6. You can chit-chat with the recommender to make the conversation more natural, brief, and fluent. \\ 7. Your utterances need to strictly follow your Seeker persona. Vary your wording and avoid repeating yourself verbatim!\\ \\ Conversation History=<HISTORY>\\ Here is your feelings about the recommender's reply: <FEELING>\\ \\ Pretend to be the Seeker! What do you say next.\\ Keep your response brief. Use casual language and vary your wording.\\ Make sure your response matches your Seeker persona, your preferred attributes, and your conversation context.\\ Do not include your feelings into the response to the Seeker!\\ Respond in the first person voice (use "I" instead of "Seeker", use "you" instead of "recommender") and speaking style of the Seeker.\end{tabular}                                                                                 \\ \bottomrule
\end{tabular}}
\caption{Prompts used in User Simulator.}
\label{tab:simulator}
\end{table*}

\begin{table*}[]
\resizebox{1\textwidth}{!}{
\begin{tabular}{ll}
\toprule
\textbf{Functions}                                           & \textbf{Prompts}                                                                                                                                                                                                                                                                                                                                                                                                                                                                                                                                                                                                                                                                                                                                                                                                                                                                                                                                                                                                                                                                                                                                                                                                                                                                                                                                                                                                                                                                                                                                                                                                                                                                                                                                                                                                                                                                                                                                                                                                                                                                                                                                                                                                                                                                                                                                                                                                                                                                                                                                                                                                                                                                                                                                                                                                                                                                                                                                                                                                                                                  \\ \hline
\begin{tabular}[c]{@{}l@{}}Watching\\ Intention\end{tabular} & \begin{tabular}[c]{@{}l@{}}You are a seeker chatting with a recommender for movie recommendation. \\ Your Seeker persona: <PROFILE>\\ Your preferred movie should cover those genres at the same time: <ATTRIBUTE\_GROUP>.\\ \\ Now you need to score your watching intention based on the criteria and recommender's utterance below:\\ \\ Watching Intention Criteria\\ \#\#\#\#\#\#\\ 1. Not Interested (Score 1): No alignment with preferred genres. Uninteresting plot and weak synopsis.\\ No favorite actors or directors involved. Poor critical acclaim. Lack of personal recommendations. \\ Inaccessible or expensive. Doesn't suit current mood or timing. Lacks originality or innovation.\\ \\ 2. Slightly Interested (Score 2): Some alignment with preferred genres, but not perfect. \\ Plot seems somewhat engaging, but not highly captivating. Some familiar faces among the cast and crew.\\ Mixed or average critical acclaim. Few personal recommendations or not strong ones. Available but may require some effort or cost.\\ Somewhat suits current mood or timing. Some degree of originality or innovation.\\ \\ 3. Moderately Interested (Score 3): Fairly good alignment with preferred genres. Intriguing plot with potential. \\ Few favorite actors or directors involved. Generally positive critical acclaim. \\ Some personal recommendations from trusted sources. Easily accessible and reasonably priced.\\ Fits current mood or timing quite well. Shows some originality or innovation.\\ \\ 4. Very Interested (Score 4): Strong alignment with preferred genres. Highly engaging plot with positive reception. \\ Many favorite actors or directors involved. High critical acclaim or praise. Several strong personal recommendations. \\ Easily accessible and included in subscription. Perfectly fits current mood or timing. Displays notable originality or innovation.\\ \\ 5. Extremely Interested (Score 5): Perfect alignment with preferred genres. Extremely captivating plot with widespread acclaim. \\ All or most favorite actors or directors involved. Exceptional critical acclaim or awards. \\ Numerous enthusiastic personal recommendations. Readily accessible and already in streaming library. \\ Perfectly suits current mood or timing. Demonstrates groundbreaking originality or innovation.\\ \#\#\#\#\#\#\\ \\ Recommender Utterance=<RECOMMENER\_EXPLANATION>\\ \\ Pretend you have little knowledge about the recommended movies, and the only information source about the movie is the recommender.\\ You can only consider your watching intention based on the information given in the recommender's utterance.\\ First summarize the movie information from the recommender utterance and consider how it matches the scoring criteria, then score your watching intention.\\ Output your reasons to the score in the "Evidence".\\ Response in the following JSON format:\\ \{"Evidence": <string>, "Watching Intention": <int>\}\end{tabular} \\ \hline
Credibility                                                  & \begin{tabular}[c]{@{}l@{}}You are an evaluator and you need to judge the credibility of the recommender's utterance based on the given source information.\\ Rate your score based on the Evaluation Standard.\\ \\ Evaluation Standard\\ \#\#\#\#\#\#\#\\ Score 5: Exact Match: The sentence exactly matches the information provided and every claim in the sentence is well-supported by the source information.\\ \\ Score 4: Supported Inference: Most of the claims in the sentence is supported by the source information with minor details can be logically inferred from the information provided.\\ \\ Score 3: Partial Match: The sentence partially matches the information provided, with some details supported and others not mentioned.\\ \\ Score 2: Inconsistent Detail: The sentence contains much information that are not supported by the information provided.\\ \\ Score 1: Unsubstantiated Claim: The sentence makes a claim that is contradict to the information provided.\\ \#\#\#\#\#\#\#\\ \\ Recommender Utterance=<RECOMMENDER\_EXPLANATION>\\ \\ Source Information=<ITEM\_INFORMATION>\\ \\ First summarize the information in the recommender' utterance and compare it with the source information to judge its credibility, then give your integer score.\\ Output your reasoning process in the "Evidence".\\  Output your score in the "Credibility".\\ \\ Response in the following JSON format:\\ \{"Evidence": <string>, "Credibility": <int>\}\end{tabular}                                                                                                                                                                                                                                                                                                                                                                                                                                                                                                                                                                                                                                                                                                                                                                                                                                                                                                                                                                                                                                                                                                                                                                                                                                                                                                                                                                                                                                                                                                                                                                          \\ \bottomrule
\end{tabular}}
\caption{Prompts used in LLM-based evaluators.}
\label{tab:evaluator}
\end{table*}

\end{document}